%% file: main_ack.tex
\documentclass{article}

\usepackage{microtype}
\usepackage{graphicx}
\usepackage{subcaption}
\usepackage{booktabs} %
\usepackage{hyperref}

\usepackage[preprint]{icml2026}

\input{sec/preamble}

\usepackage[capitalize,noabbrev]{cleveref}

\theoremstyle{plain}
\newtheorem{theorem}{Theorem}[section]
\newtheorem{proposition}[theorem]{Proposition}
\newtheorem{lemma}[theorem]{Lemma}

\theoremstyle{definition}
\newtheorem{definition}[theorem]{Definition}

\theoremstyle{remark}

\usepackage[textsize=tiny]{todonotes}

\newcommand{\jepaname}{VISReg\xspace}
\newcommand{\methodname}{VISReg\xspace}

\icmltitlerunning{\jepaname: a scaling friendly method with a better generalizability.}

\begin{document}

\twocolumn[
  \icmltitle{\jepaname: Variance-Invariance-Sketching Regularization for JEPA training}

  \icmlsetsymbol{equal}{*}

    \begin{icmlauthorlist}
    \icmlauthor{Haiyu Wu}{comp}
    \icmlauthor{Randall Balestriero}{sch1}
    \icmlauthor{Morgan Levine}{comp}
  \end{icmlauthorlist}

  \icmlaffiliation{comp}{Altos Labs}
  \icmlaffiliation{sch1}{Brown University}

  \icmlcorrespondingauthor{Haiyu Wu}{hwu@altoslabs.com}

  \icmlkeywords{Machine Learning, ICML}

  \vskip 0.3in
  \input{figures/teaser}
]

\printAffiliationsAndNotice{}  %

\input{sec/new_abstract}
\input{sec/new_intro}
\input{sec/experiment}

\input{sec/conclusion}

\section{Acknowledgment}
We appreciate Prof. Yann LeCun's efforts in connecting resources and people to bring this project to fruition.

\bibliography{citations}
\bibliographystyle{icml2026}

\newpage
\appendix
\onecolumn
\input{sec/supp}

\end{document}

%% file: sec/preamble.tex
\usepackage{amsmath}
\usepackage{amssymb}
\usepackage{mathtools}
\usepackage{amsthm}

\usepackage[most]{tcolorbox}
\usepackage{listings}
\usepackage{caption}
\usepackage{algorithm}
\usepackage{xcolor}
\usepackage{soul}
\usepackage[table]{xcolor}
\usepackage{multirow}

\usepackage[T1]{fontenc}  
\usepackage{inconsolata}  

\definecolor{algo_bg}{RGB}{245, 249, 253}
\definecolor{algo_comment}{RGB}{148, 163, 184}
\definecolor{algo_keyword}{RGB}{0, 119, 181}
\definecolor{algo_string}{RGB}{214, 95, 26}

\newcommand{\red}[1]{\textcolor{red}{#1}}
\newcommand{\gray}[1]{\textcolor{gray}{#1}}

\usepackage{xspace}

\usepackage{pifont}
\newcommand{\cmark}{\ding{51}}
\newcommand{\xmark}{\ding{55}}

\DeclareCaptionFormat{custom}{#1#2 #3}
\lstdefinestyle{screenshot_style}{
    language=Python,
    basicstyle=\ttfamily\small\color{black!80}, 
    keywordstyle=\color{algo_keyword}\bfseries,
    commentstyle=\itshape\color{algo_comment},
    stringstyle=\color{algo_string},
    showstringspaces=false,
    breaklines=true,
    frame=none,
    tabsize=4,         
    columns=fullflexible, 
    keepspaces=true 
}

\newtcolorbox{algoblock}{
    colback=algo_bg,
    colframe=white,
    sharp corners,
    boxrule=0pt,
    enhanced,
    left=5pt, right=5pt, top=5pt, bottom=5pt
}

%% file: figures/teaser.tex
\begin{center}
    \centering
    \includegraphics[width=1.0\textwidth]{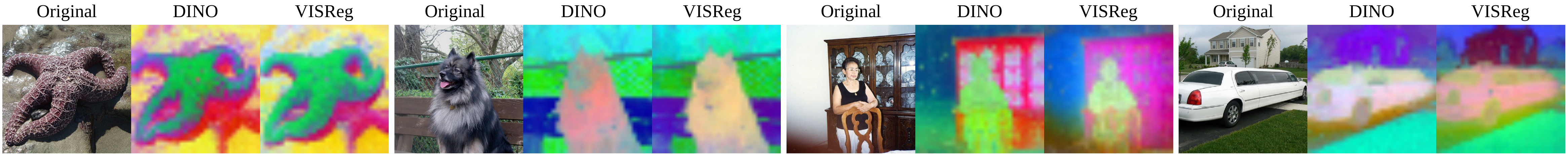} 
    \captionof{figure}{PCA visualization of last layer features. For each image, we show visualizations of features from DINO (\textbf{middle}) and \jepaname (\textbf{right}). Both methods are pre-trained on ImageNet1K with ViT-B/16. \jepaname excels in granular details than DINO without relying on any heuristics for training stability. This brings a better out-of-domain (OOD) performance and transfer learning capability.}
    \label{fig:teaser}
\end{center}

%% file: sec/new_abstract.tex
\begin{abstract}
Self-supervised learning methods prevent embedding collapse via modeling heuristics or explicit regularization of the embedding space. Among the latter, VICReg decomposes regularization into variance and covariance objectives, offering flexibility and interpretability. However, covariance captures only second-order statistics---encouraging decorrelation but failing to enforce the full distributional shape needed for stable training. Sketching-based methods such as SIGReg address this by aligning embeddings to an isotropic Gaussian, but lack flexibility and suffer from vanishing gradients under collapse. We propose Variance-Invariance-Sketching Regularization (VISReg), which replaces covariance with a Sliced-Wasserstein-based sketching objective that enforces full distributional shape, while retaining a variance term for scale control. By decoupling scale and shape, VISReg combines VICReg's flexibility with the distributional rigor of sketching methods, providing robust gradients even under collapse. We show that VISReg scales linearly, outperforms existing regularization on low-quality datasets, and is resilient to long-tailed and low-rank regimes. Pre-trained on ImageNet-1K, VISReg achieves state-of-the-art performance on out-of-distribution datasets. Pre-trained on ImageNet-22K, it matches DINOv2's OOD performance despite the latter using 10$\times$ more data (LVD-142M). Project and code: \url{https://haiyuwu.github.io/visreg}.
\end{abstract}

%% file: sec/new_intro.tex
\section{Introduction}

Self-supervised learning (SSL) has evolved from contrastive learning~\cite{simclr, simclrv2, mocov1, mocov2, mocov3} to Joint-Embedding Predictive Architectures~\cite{jepa, ijepa, dino, ibot, dinov2, dinov3}, which are more scalable and achieve stronger performance. Despite these advantages, many methods rely on heavy heuristics (\emph{e.g.,} EMA, frozen layers, teacher-student architectures) to ensure training stability.

To remove such heuristics, VICReg~\cite{vicreg} decomposes the training objective into variance, invariance, and covariance optimization. This approach largely reduces the engineering burden while achieving competitive performance. More recently, LeJEPA~\cite{lejepa} proved that sketching the embedding space toward an isotropic Gaussian is an effective principle for ensuring training stability and strong downstream performance, and proposed SIGReg based on the Epps-Pulley test~\cite{epps-pulley} and the Cram\'er-Wold theorem~\cite{cramer-wold} to realize this.

However, both methods have clear limitations. VICReg regularizes covariance, which captures only second-order statistics. While this encourages decorrelation, it cannot enforce the full distributional shape of the embedding space---a distribution can match in mean and covariance yet remain far from Gaussian. This makes covariance regularization a comparatively weak proxy for the isotropy that stable, information-rich training requires. On the other hand, SIGReg addresses distributional shape directly through sketching, but it does not decouple scale from shape, limiting flexibility across training regimes. More critically, the gradient of the Epps-Pulley test diminishes as the embedding collapses (Figure~\ref{fig:collapse}), eventually vanishing entirely---precisely when a strong corrective signal is needed most.

Motivated by these complementary shortcomings, we propose \textbf{V}ariance-\textbf{I}nvariance-\textbf{S}ketching \textbf{Reg}ularization (\textbf{VISReg}). VISReg retains the variance term from VICReg to control the scale of the embedding space, but \emph{replaces covariance regularization with sketching regularization}: we use the Sliced Wasserstein Distance (SWD)~\cite{swd} to align the normalized embedding distribution with an isotropic Gaussian prior along random 1D projections, thereby enforcing the full distributional shape. By decoupling scale and shape into separate objectives, VISReg inherits the interpretability and flexibility of VICReg's decomposed losses while leveraging the distributional rigor of sketching-based methods---and provides a robust gradient signal even under collapse. Combined with a standard invariance loss, VISReg forms a complete, heuristic-free self-supervised learning method.

We compare VISReg with SIGReg, VICReg, and DINO on both standard and low-quality datasets. We find that DINO struggles to learn meaningful embeddings without careful hyperparameter tuning, while VISReg, SIGReg, and VICReg are all robust---but VISReg achieves the highest accuracy and the most stable training, particularly on low-rank and long-tailed datasets. Our hyperparameter analyses further provide clear guidance for methods grounded in the Cram\'er-Wold theorem.

We evaluate VISReg on linear classification, transfer learning, dense prediction, and image generation guidance, covering both in-domain and out-of-distribution (OOD) settings. We pretrain backbones on ImageNet-1K and evaluate on downstream datasets. First, despite a linear probe accuracy gap relative to the best method on in-domain data, VISReg achieves the best OOD results---one of the most important properties of a useful foundation model. Second, VISReg outperforms DINO~\cite{dino} with the same backbone after fine-tuning on both in-domain and OOD datasets, even though DINO has over 3\% higher linear probe accuracy on in-domain data, indicating strong transfer learning capability. Third, a linear segmentation experiment shows VISReg performs on par with DINO for dense prediction, though a gap to the best models (\emph{e.g.,} MoCoV3~\cite{mocov3}, iBOT~\cite{ibot}) remains. Finally, to test scaling, we pretrain ViT-L/14 on ImageNet-22K~\cite{inet22k}. VISReg achieves results comparable to DINOv2~\cite{dinov2} on OOD datasets, despite the latter being trained on a 10$\times$ larger dataset (LVD-142M), demonstrating the strong potential of the VISReg approach.

The contributions of this work are:
\begin{itemize}
    \item We propose VISReg, which replaces the covariance regularization of VICReg with a sketching objective grounded in optimal transport, achieving stronger distributional control, better training stability, and resilience to low-quality datasets.
    \item We comprehensively analyze the hyperparameter landscape of VISReg and related Cram\'er-Wold-based methods, providing clear guidance for scaling and training stability within this paradigm.
    \item We demonstrate that VISReg's embedding regularization yields superior OOD generalization and strong downstream task performance, broadening the practical utility of self-supervised foundation models.
\end{itemize}

\section{Related Work}

\noindent\textbf{Contrastive Learning and Sampling Strategies.}
Early successes in self-supervised learning relied heavily on contrastive objectives, which maximize the similarity between positive pairs while pushing apart negative samples~\cite{simclr, mocov1, pirl}.
SimCLR~\cite{simclr, simclrv2} demonstrated the importance of strong data augmentation and large batch sizes.
To decouple batch size dependency, MoCo~\cite{mocov1, mocov2} introduced a momentum queue to maintain a dynamic dictionary of negative samples.
SwAV~\cite{swav} reformulated contrastive learning as an online clustering problem via the Sinkhorn-Knopp algorithm.
However, these methods rely on negative pairs or prototypes, introducing sampling bias~\cite{debiased-constrastive} and computational overhead for hard-negative mining~\cite{hard-negative-mining}. Like other non-contrastive methods, VISReg eliminates the need for negative sampling entirely.

\noindent\textbf{Masked Image Modeling (MIM).}
Inspired by BERT~\cite{bert} in NLP, MIM approaches learn by reconstructing masked inputs.
MAE~\cite{mae} and SimMIM~\cite{simmim} operate on pixel-level reconstruction, demonstrating high scalability for fine-tuning tasks.
BEiT~\cite{beit} proposes predicting discrete visual tokens.
MaskFeat~\cite{maskfeat} reconstructs HOG features to focus on structural information.
Despite excelling in transfer learning, MIM methods typically learn lower-level spatial statistics and lag behind joint-embedding methods in linear probing due to weaker semantic linear separability~\cite{linear_probe_analysis, data2vec}.

\noindent\textbf{Asymmetric Joint-Embedding Architectures.}
To avoid collapse without negatives, several methods introduce architectural asymmetry.
BYOL~\cite{byol} and SimSiam~\cite{simsiam} rely on stop-gradient operations and predictor networks to break symmetry.
Mean Teacher~\cite{mean-teachers} and DINO~\cite{dino, dinov2, dinov3} utilize a momentum-updated teacher to stabilize training, with DINO further employing centering and sharpening.
OBoW~\cite{obow} and MSN~\cite{msn} leverage prototype-based learning with asymmetric updates.
Although effective, these methods rely on implicit regularization heuristics, making their non-collapse dynamics theoretically opaque~\cite{understanding_collapse}.

\noindent\textbf{Geometric and Information-Theoretic Regularization.}
VISReg is most closely related to methods that explicitly regularize the statistical properties of embeddings.
Barlow Twins~\cite{barlow-twins} minimizes redundancy in the cross-correlation matrix between twin networks.
W-MSE~\cite{w-mse} projects embeddings onto the unit sphere and performs whitening.
VICReg~\cite{vicreg} explicitly constrains the variance, invariance, and covariance of embeddings to maximize information content. However, covariance regularization captures only second-order statistics---it encourages decorrelation but cannot enforce the full distributional shape of the embedding space. Moreover, methods like Barlow Twins and VICReg require computing covariance matrices, scaling quadratically as $\mathcal{O}(D^2)$ with embedding dimension $D$.

LeJEPA~\cite{lejepa} proved that regularizing the embedding space toward an isotropic Gaussian distribution can maintain stable heuristic-free training, and introduced SIGReg---grounded in the Epps-Pulley test~\cite{epps-pulley}---to achieve this. While SIGReg provides stronger distributional control than covariance regularization and scales linearly as $\mathcal{O}(D)$, its regulation signal diminishes when the embedding collapses. KerJEPA~\cite{kerjepa} leverages MMD to estimate the regulation of infinite projections, but incurs $O(N^2)$ complexity in batch size $N$. A contemporary work, LpJEPA~\cite{lpjepa}, proposes Rectified Distribution Matching Regularization (RDMReg) to enforce embedding sparsity.

Our VISReg bridges VICReg and SIGReg: it retains VICReg's variance term for scale control but replaces the covariance term with a sketching objective based on SWD~\cite{swd}, achieving full distributional shape regularization with robust gradients, decoupled scale-shape optimization, and linear complexity---making it well suited for scaling.

\section{VISReg: Variance-Invariance-Sketching Regularization}
\label{sec:visreg}

VICReg~\cite{vicreg} decomposes embedding regularization into variance and covariance terms, providing interpretability and flexibility. However, covariance regularization captures only second-order statistics: it encourages decorrelation among embedding dimensions but cannot enforce the full distributional shape of the embedding space. A distribution can match in mean and covariance yet remain far from the isotropic Gaussian that LeJEPA~\cite{lejepa} proved to be optimal for stable, heuristic-free self-supervised training.

SIGReg~\cite{lejepa} addresses this by directly sketching the embedding distribution toward an isotropic Gaussian via the Epps-Pulley test and the Cram\'er-Wold theorem. However, we identify two limitations: (1) its gradient diminishes as the embedding collapses (Figure~\ref{fig:collapse}), vanishing precisely when correction is needed most; and (2) it does not decouple scale from shape, limiting flexibility across training regimes.

VISReg resolves these limitations by replacing VICReg's covariance term with a sketching objective while retaining the variance term for scale control. By decoupling regularization into distinct \emph{scale} and \emph{shape} objectives, VISReg provides robust gradients against collapse, distributional rigor beyond second-order statistics, and the flexibility to reweight objectives for different data regimes.

\subsection{Regularization Loss}

We decouple the regularization into scale and shape components, each operating independently. For simplicity, the number of augmentations $V$ is omitted from the derivation.

\textbf{Scale Regularization.}
We regulate the scale of the embedding space using a variance constraint, following the same intuition as VICReg. Directly minimizing the KL divergence~\cite{kl-loss} to an isotropic Gaussian prior incurs $O(D^3)$ complexity. We relax this by factorizing into marginal distributions. Given the centered embedding $\mathbf{\hat{Z}} \in \mathbb{R}^{N \times D}$, the scale loss is:
\begin{equation}
    \mathcal{L}_{\mathrm{scale}} = \frac{1}{D} \sum_{j=1}^{D} (1 - \sigma_j(\mathbf{\hat{Z}}))^2
\end{equation}
where $\sigma_j(\cdot)$ denotes the standard deviation of the $j$-th dimension. This formulation provides a gradient that approaches a constant during collapse, ensuring a reliable corrective signal.

\input{figures/collapse-prevention}

\textbf{Shape Regularization.}
Where VICReg uses covariance to encourage decorrelation---capturing only second-order statistics---we instead \emph{sketch} the embedding distribution toward an isotropic Gaussian, enforcing full distributional shape. To isolate the geometric structure from magnitude, we normalize $\mathbf{\hat{Z}}$:
\begin{equation}
    \mathbf{\widetilde{Z}} = \frac{\mathbf{\hat{Z}}}{sg(\sigma) + \epsilon}
\end{equation}
The stop-gradient $sg(\cdot)$ decouples shape optimization from scale, ensuring gradients from the shape loss do not interfere with variance regulation. Unlike prior uses of stop-gradient as a collapse-prevention heuristic~\cite{byol, simsiam}, here it serves a principled role in objective decomposition.

To efficiently align the high-dimensional distribution of $\mathbf{\widetilde{Z}}$ with the isotropic Gaussian prior, we leverage the \textit{Sliced Wasserstein Distance}, grounded in the Cram\'er-Wold theorem~\cite{cramer-wold}:

\begin{lemma}[Cram\'er-Wold Theorem]
\label{lemma:cramer_wold}
Let $\mu$ and $\nu$ be two probability measures on $\mathbb{R}^d$. The Radon transform~\cite{radon-transform} $\mathcal{R}$, defined as $\mathcal{R}\mu(\theta, t) := \int_{\mathbb{R}^d} \delta(t - \langle x, \theta \rangle)\, d\mu(x)$ along all directions $\theta \in \mathbb{S}^{d-1}$, is injective. Thus:
\begin{equation}
    \mu = \nu \iff \mathcal{R}\mu(\theta, \cdot) = \mathcal{R}\nu(\theta, \cdot), \quad \forall \theta \in \mathbb{S}^{d-1}.
\end{equation}
\end{lemma}

This allows us to regularize the high-dimensional shape by aligning 1D random projections $P_k = \mathbf{\widetilde{Z}} w_k$, where $w_k \in \mathbb{R}^{D}$. Unlike SIGReg, which operates in the frequency domain via the Epps-Pulley test, we adopt the \textbf{2-Wasserstein distance ($\mathcal{W}_2$)}, which admits an efficient closed-form solution in 1D~\cite{1d-w2-close-form-solution, swd, swd-gan}:

\begin{lemma}[1D Wasserstein Closed-Form]
\label{lemma:1d_wasserstein}
For one-dimensional distributions, the $p$-th Wasserstein distance equals the $L_p$ distance between quantile functions. For discrete empirical samples of size $N$:
\begin{equation}
    \mathcal{W}_p^p(\hat{\mu}, \hat{\nu}) = \frac{1}{N} \sum_{i=1}^{N} \| x_{(i)} - y_{(i)} \|^p,
\end{equation}
where $x_{(i)}$ denotes the $i$-th order statistic.
\end{lemma}

Leveraging Lemma~\ref{lemma:1d_wasserstein} with $p=2$, the shape loss is:
\begin{equation}
    \mathcal{L}_{\mathrm{shape}} = \frac{1}{K} \sum_{k=1}^{K} \left\| \mathrm{sort}(\mathbf{\widetilde{Z}} w_k) - \mathbf{q}_{\mathcal{N}} \right\|_2^2,
\end{equation}
where $\mathrm{sort}(\cdot)$ sorts the projected values in each direction, and $\mathbf{q}_{\mathcal{N}} \in \mathbb{R}^{N}$ represents the fixed quantiles of the standard Gaussian distribution. This is strictly more expressive than covariance regularization: it enforces not just decorrelation but the full marginal distribution along every projected direction.

Additionally, empirical results suggest that regularizing the embedding center increases training robustness, so we include a centering loss:
\begin{equation}
    \mathcal{L}_{\mathrm{center}} = \|\mu\|_2^2
\end{equation}
where $\mu$ is the batch mean.

\begin{proposition}[VISReg Regularization Objective]
The regularization loss $\mathcal{L}_{\mathrm{Reg}}$ optimizes variance and distributional shape independently:
\begin{equation}
    \mathcal{L}_{\mathrm{Reg}} = \lambda_{\mathrm{scale}}\, \mathcal{L}_{\mathrm{scale}} + \lambda_{\mathrm{shape}}\, \mathcal{L}_{\mathrm{shape}} + \lambda_{\mathrm{center}}\, \mathcal{L}_{\mathrm{center}}
\end{equation}
\end{proposition}

The code is shown in Algorithm~\ref{alg:visreg}. Decoupling introduces three hyperparameters; we conduct ablations in Table~\ref{tab:dsso-ratio-ablation}. \textit{The default $\lambda_* = 1$ works well for high-quality datasets, but increasing the shape loss weight improves performance on low-quality datasets.}

\input{code/visreg}

For the invariance objective, we follow LeJEPA~\cite{lejepa}:
\begin{equation}
    \mathcal{L}_{\mathrm{pred}} = \frac{1}{V} \sum_{i=1}^{V} \|\mu_g - z_i\|_2^2
\end{equation}
where $V$ is the number of views, $\mu_g$ is the mean embedding of global views, and $z_i$ includes both global and local view embeddings. The full VISReg objective is:
\begin{equation}
    \mathcal{L}_{\mathrm{VISReg}} = (1 - \lambda)\, \mathcal{L}_{\mathrm{pred}} + \lambda\, \mathcal{L}_{\mathrm{Reg}}
\end{equation}

Ablation results for each component are in Section~\ref{ablation:visreg}.

\subsection{VISReg Is Friendly to Scale Up}
\label{sec:kd-correlation}

One practical advantage inherited from the Cram\'er-Wold framework is favorable scaling behavior. Due to limited computational resources, we analyze scalability through algorithm complexity, simulated scaling cost, and experiments on a small yet challenging dataset.

\begin{definition}
Let the input feature $\mathbf{Z} \in \mathbb{R}^{N \times D}$, where $N$ is the mini-batch size and $D$ is the projection dimension. The number of random slices is $K$.
\end{definition}

The complexity of $\mathcal{L}_{\mathrm{Reg}}$ is dominated by two operations:
\begin{equation}
    \mathcal{C}_{\mathrm{Reg}} = \underbrace{O(NDK)}_{\text{projection}} + \underbrace{O(KN\log N)}_{\text{sorting}}
\end{equation}
Since $\log N \ll D$ at scale, the effective complexity is:
\begin{equation}
    \mathcal{C}_{\mathrm{Reg}} = O(NDK)
\end{equation}
This is linear in all scaling parameters---compared to VICReg's $O(ND^2)$ from covariance computation. We next analyze the effects of batch size $N$, projection dimension $D$, and number of slices $K$.

\input{figures/scaling_cost}

\textbf{Analysis in $N$.} We simulate the running time and memory demand of popular regularization methods~\cite{barlow-twins, vicreg, lejepa} at scale. Since VISReg is based on SWD, we also include vanilla SWD. Figure~\ref{fig:scale} shows that SWD-based methods are more efficient in both speed and memory. The 17-knot sampling required by the Epps-Pulley test slows SIGReg down. \textit{We conclude that VISReg scales efficiently in batch size.}

Lemma~\ref{lemma:cramer_wold} establishes that we can regulate $D$-dimensional space by aligning $K$ 1D slices, so the relationship between $K$ and $D$ is important for scaling. We analyze this by reporting the online linear probe accuracy of ViT-S/8~\cite{vit} on ImageNette\footnote{https://github.com/fastai/imagenette}~\cite{imagenet}, comparing SIGReg~\cite{lejepa} (CF-based) with SWD~\cite{swd} and VISReg (OT-based). Unless stated otherwise, we use a batch size of 256, a learning rate of $10^{-3}$ without decay, 4 global views with a cropping ratio (0.08, 1). The $\lambda$ weights for SIGReg, SWD, and VISReg are 0.02, 0.6, and 0.6, respectively. Models are trained on a single H100 GPU for 800 epochs; we report the highest accuracy.

\input{figures/proj_dim}
\textbf{Analysis in $D$.} With a sufficient number of slices ($K = 4096$), we vary the projection dimension $D$. Figure~\ref{fig:proj_dim} reveals three patterns: (1) OT-based methods regularize dimensions more efficiently than the CF-based method; (2) with sufficient $K$, VISReg learns better semantically meaningful embeddings; (3) $K$ must exceed $D$ by a factor $C > 1$ for optimal accuracy, and VISReg requires the smallest $C$. The third observation suggests that $K$ cannot be treated as independent of $D$, converting the effective complexity from $O(NDK)$ to $O(ND \cdot CD)$. We address this below.

\input{figures/num_slices}
\textbf{Analysis in $K$.} Fixing $D = 256$ and varying $K$, Figure~\ref{fig:num-slices} shows: (1) OT-based methods remain robust even at $K = \frac{1}{8}D$; (2) VISReg is the most robust approach, consistently achieving the highest linear probe accuracy.

\input{figures/gpu-effect}
Despite VISReg's robustness, the correlation between $K$ and $D$ remains a concern for complexity. Revisiting Lemma~\ref{lemma:1d_wasserstein} and Algorithm~\ref{alg:visreg}, we observe that $K$ random slices are generated independently per GPU, so one can generate $\frac{CD}{M}$ slices on each of $M$ GPUs to obtain $K = CD$ total slices. For example, 128 slices per GPU on 8 GPUs should match 1024 slices on one GPU.

Figure~\ref{fig:gpu-effect} confirms this. With one GPU, the accuracy gap between $K = 128$ and $K = 1024$ reaches 13.88\% for SIGReg, 2.21\% for SWD, and 2.44\% for VISReg. With 8 GPUs and the same per-GPU $K$, the gap shrinks to 0.27\%, 0.24\%, and 0.22\% respectively. Given nondeterministic training, these results support our claim. \textit{Thus, $K$ can remain constant when scaling, preserving the $O(NDK)$ complexity.}

\subsection{VISReg Is Robust to Low-Quality Datasets}

Low-quality datasets pose challenges from many angles. We evaluate on ImageNet-LT~\cite{imagenet-lt} (long-tailed) and Galaxy10~\cite{galaxy10} (low-rank). Training settings follow the previous section except $lr = 10^{-4}$, $K = 4096$, $D = 256$, with images resized to 128px. We include DINO~\cite{dino} and VICReg as baselines. Models are trained from scratch for 400 epochs to simulate real-world scenarios where suitable pretrained models do not exist---a common challenge in domains like AI for Science.

\input{tables/imagenet-lt}
ImageNet-LT is a long-tailed variant of ImageNet-1K containing 115K images from 1K classes, categorized into many-shot, medium-shot, and few-shot. Table~\ref{tab:imagenet_lt_results} shows that VISReg outperforms all methods at all levels after adjusting the shape loss weight (details in Table~\ref{tab:dsso-ratio-ablation}), whereas DINO fails to learn meaningful embeddings.

\input{tables/galaxy10-raw}
Galaxy10 comprises 17,736 galaxy images from 10 classes. We treat it as low-rank because: (1) it has 10 classes with limited training data, below the capacity of ViT-S/8; and (2) most images contain a large ratio of black pixels, limiting useful content. Table~\ref{tab:galaxy10-raw} shows that all four regularization methods prevent collapse and achieve good accuracy, but DINO struggles to learn meaningful embeddings.

\textbf{Summary of Analyses.} First, VISReg has complexity $O(NDK)$, linear in all scaling factors---an improvement over VICReg's $O(ND^2)$. Second, we observe that $K$ is correlated with $D$ by a factor $C > 1$, but prove that distributing slices across $M$ GPUs resolves this, keeping $K$ constant at scale. Third, VISReg outperforms existing methods in training efficiency, effectiveness, and robustness. Fourth, VISReg is more resilient to low-quality datasets through loss reweighting---demonstrating the importance of decoupling scale and shape over the monolithic covariance or sketching approaches. All these results confirm that VISReg is a practical and principled regularization method for real-world self-supervised learning.

\input{tables/ablation-study}

%% file: figures/collapse-prevention.tex
\begin{figure}[t]
    \centering
    \includegraphics[width=0.8\linewidth]{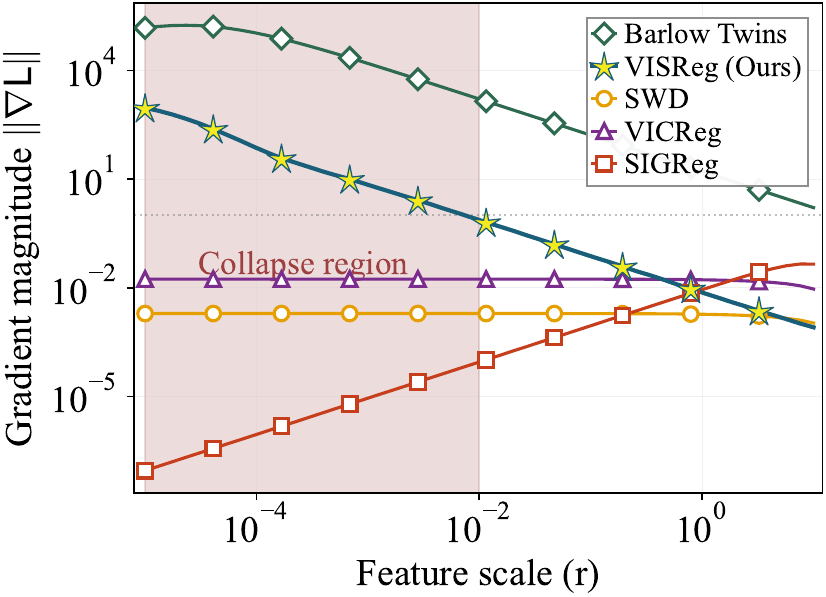}
    \caption{Embedding collapse prevention. We simulate the gradient $\left|\left|\nabla L\right|\right|$ of popular regularization methods under different collapse stages by changing the feature norm ($r$). We observe that when the model is collapsed, Barlow Twins~\cite{barlow-twins} and \methodname provide a strong gradient to fix the collapse, whereas SIGReg~\cite{lejepa} fails to do so.}
    \label{fig:collapse}
\end{figure}

%% file: code/visreg.tex
\begin{algorithm}[t]
    \caption{Decoupled regularization term in \methodname. $z$ is a $(N, D)$ tensor, $K$ is the number of slices.}
    \label{alg:visreg} 
    \begin{algoblock}
    \begin{lstlisting}[style=screenshot_style]
def visreg(z, K=64):
    # 1. Center loss
    mu = z.mean(dim=0)
    L_center = (mu).pow(2).mean()
    # 2. Scale loss
    z_cent = z - mu
    std = z_cent.std(dim=0, unbiased=False)
    L_scale = (1.0 - std).pow(2).mean()

    # 3. Shape loss: SWD
    z_norm = z_cent / (std.detach())
    W = torch.randn(D, K)
    W /= W.norm(p=2, dim=0)
    
    # Project and sort
    p = z_norm @ W
    p_sorted = torch.sort(p, dim=0).values
    u = torch.arange(1, N+1) / (N+1)
    target = Normal(0, 1).icdf(u)
    
    L_shape = (p_sorted - target).pow(2).mean()
    
    return L_scale + L_shape + L_center
    \end{lstlisting}
    \end{algoblock}
\end{algorithm}

%% file: figures/scaling_cost.tex
\begin{figure}
    \centering
    \includegraphics[width=\linewidth]{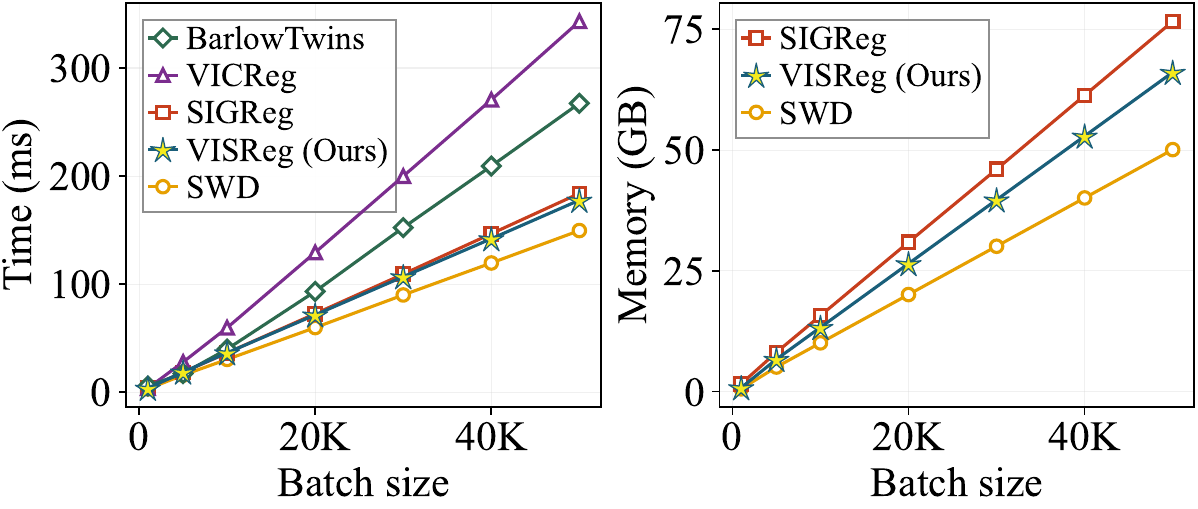}
    \caption{Scaling cost. We simulate the cost of popular regularization methods after scaling the model at different batch sizes. On a single H100 (80GB) GPU, our method achieves a slightly better speedup with a 13.7\% memory demand over SIGReg at a batch size of 50K. The projection dimension, number of slices, and the number of views are 10K, 2.5K, and 8.
    }
    \label{fig:scale}
\end{figure}

%% file: figures/proj_dim.tex
\begin{figure}[t]
    \centering
    \includegraphics[width=\linewidth]{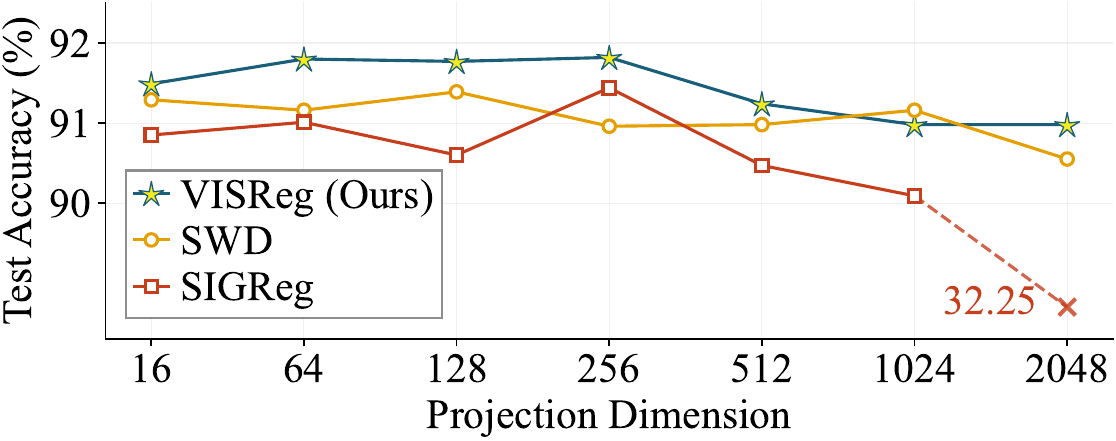}
    \caption{Linear probe accuracy with different projection dimensions ($D$). We vary $D$ with a fixed number of slices ($K=4096$) on three Cram\'er-Wold-based methods. It indicates that $K$ must be larger than $D$ by a factor of $C > 1$ to maintain the best accuracy, so these approaches are O(CD$^2$) to scaling factors on one GPU.
    }
    \label{fig:proj_dim}
\end{figure}

%% file: figures/num_slices.tex
\begin{figure}[t]
    \centering
    \includegraphics[width=\linewidth]{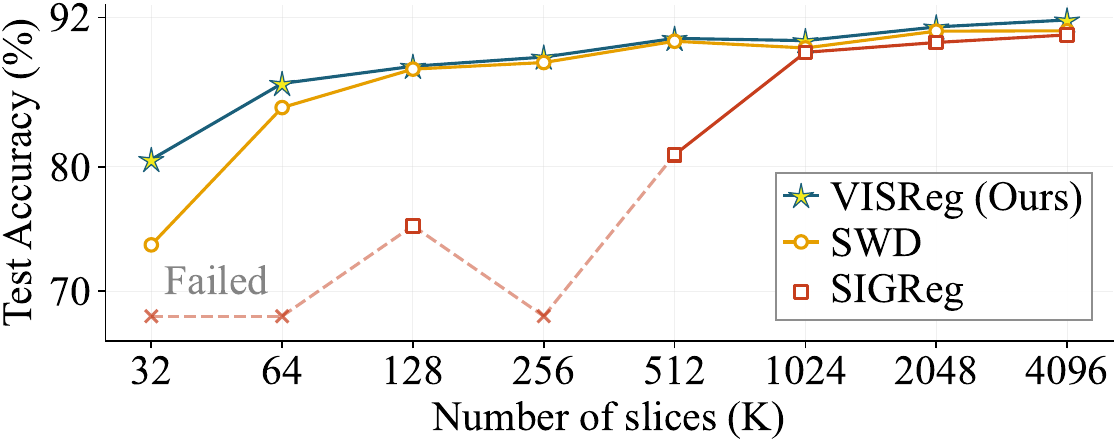}
    \caption{Linear probe accuracy with different numbers of 1D slices ($K$). The projection dimension $D$ is 256 and $K$ varies from $\frac{1}{8}D$ to $16D$. It shows that DSSO is robust even with $K=\frac{1}{8}D$.}
    \label{fig:num-slices}
    \vspace{-3mm}
\end{figure}

%% file: figures/gpu-effect.tex
\begin{figure}[t]
    \centering
    \includegraphics[width=\linewidth]{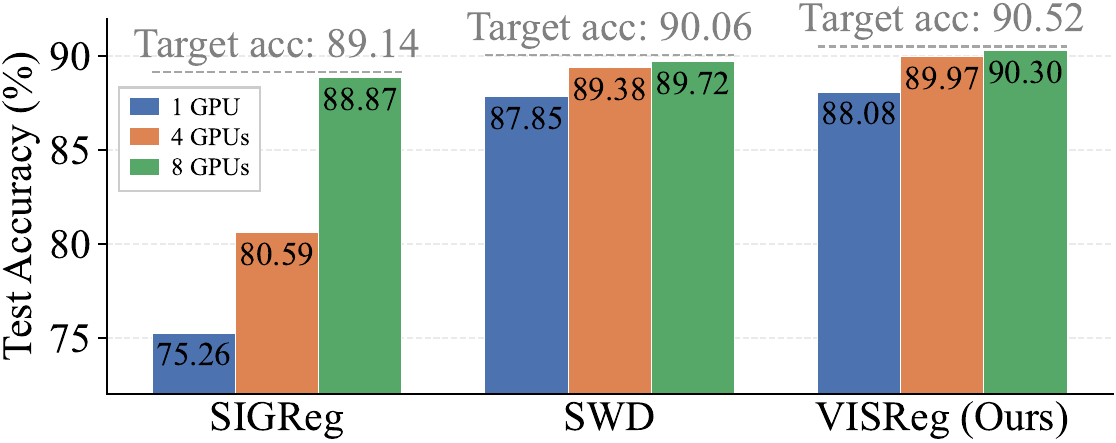}
    \caption{Linear probe accuracy in scaling the number of GPUs with the fixed $K$ and $D$. This result indicates that scaling the number of GPUs can compensate for the insufficient $K$=$\frac{1}{4}D$ to a sufficient level. When using 8x more GPUs, the final accuracy matches the target accuracy of $K$=$2D$, which makes $K$ a constant number possible when scaling the training.}
    \label{fig:gpu-effect}
    \vspace{-5mm}
\end{figure}

%% file: tables/imagenet-lt.tex
\setlength{\tabcolsep}{3.4mm}
\begin{table}[t]
\centering
\caption{Linear probe accuracy on ImageNet-LT. The backbone, ViT-S/8, is trained for 400 epochs from scratch. Our \methodname method outperforms all methods at all levels. DINO fails to learn meaningful embeddings. The accuracy values are reported in percentage. $*$ means increasing the weight of shape loss.}
\label{tab:imagenet_lt_results}
\begin{tabular}{l|cccc}
\hline
Method & Overall & Many & Medium & Few \\ \hline
SWD    & 31.85 & 51.54 & 22.70 & 8.36 \\
SIGReg & 32.00 & 51.86 & 22.88 & 7.92 \\
\rowcolor{blue!15} \methodname   & 32.11 & 51.55 & 23.19 & 8.52 \\
\rowcolor{blue!15} \methodname\hspace{-1mm}$^*$   & \textbf{35.14} & \textbf{54.49} & \textbf{26.87} & \textbf{9.40} \\ \hline
VICReg    & 33.08 & 52.29 & 24.63 & 8.54 \\
DINO   & \red{5.13} & \red{12.22} & \red{0.82} & \red{0.24} \\
\hline
\end{tabular}
\end{table}

%% file: tables/galaxy10-raw.tex
\setlength{\tabcolsep}{0.7mm}
\begin{table}
    \centering
    \caption{In-domain linear probe accuracy on Galaxy10. The model is trained from scratch to test the performance of methods on the low-rank task. SIGReg, SWD, and \methodname successfully prevent the training from collapsing while obtaining a good linear probe accuracy, whereas DINO struggles to learn meaningful features. $*$ means increasing the weight of shape loss.}
    \begin{tabular}{c|cc >{\columncolor{blue!15}}c >{\columncolor{blue!15}}c |cc}
    \hline
         & SWD & SIGReg & \methodname & \methodname\hspace{-1mm}$^*$ & VICReg & DINO\\ 
        Acc. & 80.60 & 80.50 & 80.51 & \textbf{80.76} & 79.93 & \red{73.49} \\
    \hline
    \end{tabular}
    \label{tab:galaxy10-raw}
    \vspace{-3mm}
\end{table}

%% file: tables/ablation-study.tex
\setlength{\tabcolsep}{2.1mm}
\begin{table*}[t]
\caption{Ablation study of training hyper-parameters. From left to right, we conduct the ablation experiment on $\lambda$, learning rate, batch size, and projection dimension. The ViT-B/16 backbone is trained for 100 epochs for the first three tables and 300 epochs for the last one.}
\label{tab:ablation}
    \vspace{2mm} %
    
    \noindent %
    \hspace*{\fill}%
    \begin{tabular}[t]{c|c}
    \hline
        $\lambda$ & Acc. \\ \hline
        0.7 & 67.98\\
        0.8 & 68.64\\
        \rowcolor{cyan!15} \textbf{0.9} & \textbf{69.19}\\
        0.95 & 69.04\\
    \hline
    \end{tabular}%
    \hfill %
    \begin{tabular}[t]{c|c}
    \hline
        Learning rate & Acc. \\ \hline
        1e-4 & 65.73\\
        3e-4 & 68.85\\
        5e-4 & 69.19\\
        7e-4 & 69.28\\
        \rowcolor{cyan!15} \textbf{9e-4} & \textbf{69.66}\\
    \hline
    \end{tabular}%
    \hfill %
    \begin{tabular}[t]{c|c}
    \hline
        Batch size & Acc. \\ \hline
        128 & 67.63 \\
        256 & 69.43 \\
        \rowcolor{cyan!15} \textbf{512} & \textbf{69.66} \\
        1024 & 69.45 \\
        2048 & 68.79 \\
    \hline
    \end{tabular}%
    \hfill %
    \begin{tabular}[t]{c|c}
    \hline
        Projection dim. & Acc. \\ \hline
        64 & 73.20\\
        128 & 73.34\\
        256 & 73.44\\
        \rowcolor{cyan!15} \textbf{512} & \textbf{73.53}\\
    \hline
    \end{tabular}%
    \hspace*{\fill}%
\end{table*}

\setlength{\tabcolsep}{1.15mm}
\begin{table*}[t]
    \caption{Effect of projection dimension on downstream tasks. We find that there is no one-size-fit-all setting, as the optimal projection dimension varies across downstream tasks. We report the linear probe performance on seven in-domain datasets (\textbf{left}) and three OOD datasets (\textbf{middle}), and linear segmentation on ADE20K (\textbf{right}). The metric is AU-ROC for ChestXRay, mIoU for ADE20K, and accuracy for the other datasets. 
    The training epoch is 40 for ADE20K and 10 for the others. The \textbf{best} and the \underline{second best} values are highlighted.}
    \label{tab:proj_dim_down_stream}
    \centering
    \begin{tabular}{c|ccccccc|ccc|c}
    \hline
        Projection dim. & Aircrafts & Cars & Cifar10 & Cifar100 & Flowers & Food & Pets & DTD & Galaxy10 & ChestXRay & ADE20K\\ \hline
        64 & 50.70 & 60.89 & 92.43 & 75.32 & 84.34 & 82.29 & 83.88 & 71.39 & \underline{71.99} & \underline{0.7550} & \textbf{30.08} \\
        128 &  \textbf{51.51} & 60.86 & 94.18 & 77.39 & 85.38 & \textbf{82.51} & 84.17 & \textbf{72.51} & 71.37 & 0.7542 & \underline{30.06} \\
        \rowcolor{cyan!15} 256 &  50.22 & \underline{60.97} & \underline{94.22} & \underline{78.93} & \textbf{86.19} & \underline{82.44} & \underline{84.64} & 71.09 & \textbf{73.33} & \textbf{0.7561} & 29.69 \\
        512 &  \underline{51.00} & \textbf{61.55} & \textbf{94.99} & \textbf{80.38} & \underline{85.89} & 82.34 & \textbf{84.69} & \underline{71.97} & 70.93 & 0.7543 &  28.94\\
        \hline
    \end{tabular}
\end{table*}

%% file: sec/experiment.tex
\section{Experiment}
This section covers the ablation study of hyperparameter settings, the effect of projection dimension on downstream tasks, and comparisons between \jepaname and existing methods in linear probe, transfer learning, domain shifting, dense instance prediction, and image generation guidance.

\subsection{Ablation study}
Unlike previous works~\cite{dino, mocov3, ijepa} relying on heuristics for training stability, \jepaname only has four hyper-parameters to tune. Unless stated otherwise, the training set is ImageNet1K, the backbone is ViT-B/16, the number of slices is 4096 per GPU, the augmentation settings follow LeJEPA~\cite{lejepa} with 2 global views and 6 local views, and the training epoch is 100. We report the online linear probe accuracy on ImageNet1K to analyze the effect of hyper-parameters, as shown in Table~\ref{tab:ablation}.

\textbf{Effect of $\lambda$.} Different from SIGReg, scaling the regularization loss with batch size to maintain the batch size invariance, \methodname is naturally batch invariant, so a large $\lambda$ value is needed to ensure the contribution of \methodname in the gradient. 
\textit{For small datasets, e.g., ImageNette and Galaxy10, 0.6 is a good start. For large datasets, e.g., ImageNet1K, 0.9 is a good start.}

\textbf{Effect of learning rate.} Similar to the other methods, 5e-4 to 1e-3 is the optimal range for the training on ImageNet1K. \textit{When training on a large dataset, 9e-4 is a good start.}

\textbf{Effect of batch size.} Grounded in the same theorem as LeJEPA, \jepaname is also robust to a small batch size. Different from LeJEPA, the \methodname algorithm in \jepaname benefits from a large batch size in regularizing the embedding space. \textit{Hence, we recommend reducing $\lambda$ when observing a fast accuracy saturation with a large batch size.}

\textbf{Effect of projection dimension.} 
Similar to previous works, we use a 3-layer MLP as the projection layer to apply regularization. Different from previous works, the final projection dimension not only decides the information bandwidth but also the difficulty of the regularization process, \emph{i.e.,} the lower dimension the easier. To investigate the trade-off, we increase the training epochs to 300 and test the model performance under four projection dimension settings.

The investigation includes three aspects: in-domain, OOD, and segmentation. Following the settings in DINOv2~\cite{dinov2}, we run offline linear probe on ImageNet1K and linear segmentation on ADE20K~\cite{ade20k}. The full-shot linear probe performance of the other datasets~\cite{aircrafts, cars, cifar, flowers, pets, dtd, chestxray} is reported for pattern observation. 

\input{tables/main_results}
Starting at the in-domain results, one observation is that a larger projection dimension results in a higher accuracy. With the embedding size 768 of ViT-B/16, projection dimension 512 gives the highest overall accuracy and 64 gives the lowest average accuracy. This indicates that projection dimension can be the bottleneck for in-domain classification.
Focusing on the OOD datasets, the observation is that a smaller projection dimension limits the OOD performance, but a larger dimension might lead to over-parameterization and training set memorization. Interestingly, the smallest the projection dimension outperforms the largest one. 
Lastly, we observe that a smaller projection dimension leads to a better performance on dense instance prediction. We choose 256 as the optimal setting in the training.

\subsection{General comparison}
Linear probe, transfer learning, and domain shifting are three key aspects of evaluating the efficacy of a SSL foundation model. We compare \jepaname with seven existing methods that are widely used in the real-world applications. In addition, we add the segmentation and generation tasks to evaluate \jepaname on dense instance prediction and semantic meaning guidance for generation.

\input{tables/lp-ood-test}
\textbf{Datasets.} The base training set is ImageNet1K~\cite{imagenet}. There are 15 datasets used to cover the general comparison experiment. 8 of them are in-domain datasets: FGVC-aircraft~\cite{aircrafts}, Stanford cars~\cite{cars}, Cifar10 \& Cifar100~\cite{cifar}, Oxford 102 flowers~\cite{flowers}, Food 101~\cite{food}, Oxford-IIIT Pet~\cite{pets}, and ImageNet1K. 6 of them are OOD datasets: Describable Textures Dataset (DTD)~\cite{dtd}, Galaxy10~\cite{galaxy10}, ChestXRay~\cite{chestxray}, Aerial Image Dataset (AID)~\cite{aid}, RetinaMNIST~\cite{retinamnist}, and OrganAMNIST~\cite{organamnist}. The last dataset is ADE20K~\cite{ade20k} for dense instance prediction. The details are in~\ref{supp/dataset}.

\textbf{Training settings.} We choose two commonly used backbones, ViT-B/16 and ViT-L/14, to run the experiments. ViT-B/16 uses the best hyperparameters in Table~\ref{tab:ablation} and Table~\ref{tab:proj_dim_down_stream}. ViT-L/14 is trained with \{learning rate=8e-4, $\lambda$=0.7, batch size=512, projection dim=384\}. Both backbones are trained for 400 epochs and 4 global + 6 local views are used. The other settings follow LeJEPA~\cite{lejepa}. We directly use the timm package to create the model and load the pre-trained weights.

\textbf{Downstream task evaluation settings.} The linear probe, transfer learning, and linear segmentation experiments use the same settings as DINOv2~\cite{dinov2}. The only difference is that the training epoch of linear probe on downstream datasets is 10. We only compare with the models that are pre-trained on ImageNet1K.

\textbf{\jepaname has a competitive in-domain performance.} Table~\ref{tab:in-domain} groups the methods based on the heuristics utilization. Within the w/o heuristics group, \jepaname has a stronger in-domain performance than MAE~\cite{mae} and LeJEPA~\cite{lejepa}, achieving 75.7\% accuracy with ViT-B/16 and 77.0\% accuracy with ViT-L/14 on ImageNet1K. Moreover, \jepaname achieves the best average accuracy on downstream datasets. Comparing to the w/ heuristics methods, there is still an accuracy gap. Despite the accuracy gap on in-domain datasets, \jepaname indicates a stronger performance on DTD, the only OOD dataset in the table. Note that the ViT-B/16 of \jepaname even outperforms the ViT-L and ViT-H of the other methods. This intriguing observation motivates the extended experiments on more OOD datasets.

\textbf{\jepaname has a better OOD performance.}
Due to the lack of OOD evaluations in previous work, we select 6 datasets from distinct domains: ChestXRay, RetinaMNIST, and OrganAMNIST are from medical domain, Galaxy10 is from space domain, and AID has the aerial images.
The results in Table~\ref{tab:lp-ood} suggest that \jepaname helps the model learn more general features than the other methods. Without using training heuristics, \jepaname achieves the best average accuracy comparing with all methods and backbone scales. Moreover, after scaling the training set to ImageNet22K, \jepaname with ViT-L/14 backbone achieves a comparable accuracy to DINOv2, which was trained with a 10x larger training set. This indicates the generality of the representations learned by \jepaname. This advantage also benefits the transfer learning capability. 

\input{tables/ft-test}
\textbf{\jepaname has a good transfer learning capability.}
We conduct a transfer learning experiment on CIFAR10 \& CIFAR100, Flowers, ImageNet1K, and Galaxy10. To have a fair comparison with DINO, the backbone is ViT-B/16 and the fine-tuning follows DINO~\cite{dino} implementation. An important observation is that, although \jepaname does not have a better linear projection accuracy on in-domain datasets than DINO, the fine-tuning results are consistently higher than both supervised learning~\cite{sup} and DINO. In addition, the advantage of \jepaname on OOD datasets still remains.

\input{tables/seg}
\textbf{\jepaname shows an on-par performance on dense instance prediction.}
Following DINOv2~\cite{dinov2}, we conduct a simple linear segmentation experiment on ADE20K and the mIoU result is reported. Table~\ref{tab:seg} indicates that, without using any heuristics, \jepaname can still provide a good segmentation results. Nevertheless, there is still a large gap comparing with MoCoV3 and iBOT, which is an important aspect that we will work on.

\input{tables/generation}

\textbf{\jepaname provides a good guidance to speed up the training of generative models.} Another important application of foundation model is to speed up the training process of generative model~\cite{repa, irepa}. We use the official code of iREPA~\cite{irepa} and run a lightweight training on SiT-B/2 for 100K steps with the features from \jepaname and DINO. We use the default settings for both training and generation. The results in Table~\ref{tab:gen} suggest that \jepaname provides useful embeddings.

%% file: tables/main_results.tex
\setlength{\tabcolsep}{1.65mm}
\begin{table*}[t]
    \centering
    \caption{Linear probe (LP) accuracy on Inet1K and downstream datasets. Comparing with the existing methods with different backbone scales, \jepaname has a competitive performance to the methods with heuristics and a better performance than the methods without heuristics. Looking at the accuracy on the \sethlcolor{red!20}\hl{OOD dataset}, \jepaname outperforms all methods that use heuristics, which suggests more general features are learned. The values at row$^*$ and col$^*$ are borrowed from the original paper.}
    \label{tab:in-domain}
    \begin{tabular}{llc >{\columncolor{red!20}}c cccccccc|c}
    \hline
        Methods & Backbone & Epochs & DTD & Aircraft & Cars & Cifar10 & Cifar100 & Flowers & Food & Pets & Avg. & Inet1K$^*$\\ \hline
        \multicolumn{3}{l}{\gray{w/ heuristics  - LP 10 epochs}} & \cellcolor{white} & & & & & & & &\\
        MoCoV3 & ViT-B/16  & 300 & 73.7 & 57.9 & 67.5 & 96.9 & 85.2 & 91.5 & 81.8 & 89.8 & 80.5 & 76.7\\
        DINO & ViT-B/16  & 400 & 74.3 & 63.6 & 73.9 & 96.5 & 85.0 & \textbf{94.6} & 83.1 & 93.6 & 83.1 & 78.2\\
        data2vec & ViT-L/14 & 1600 & 69.7 & 43.9 & 38.7 & 96.9 & 83.7 & 81.4 & 79.6 & 83.0 & 72.1 & 77.3\\
        iBOT & ViT-B/16 & 400 & 74.1 & 63.5 & 73.8 & 97.1 & 85.9 & 93.7 & 84.2 & 93.6 & 83.2 & 79.8\\
        iBOT & ViT-L/16 & 250 & 75.3 & \textbf{66.0} & \textbf{76.1} & \textbf{97.5} & \textbf{87.2} & 94.0 & \textbf{86.1} & \textbf{94.0} & \textbf{84.5} & \textbf{81.0}\\
        I-JEPA & ViT-H/14  & 300 & 69.9 & 55.4 & 59.2 & 97.2 & 85.5 & 86.8 & 83.3 & 92.8 & 78.7 & 79.3\\
        \multicolumn{3}{l}{\gray{w/o heuristics - LP 10 epochs}} & \cellcolor{white} & & & & & & & & &  \\
        MAE & ViT-L/16 & 1600  & 72.8 & 61.9 & 61.5 & 93.3 & 78.0 & 85.4 & 78.6 & 91.3 & 77.8 & 75.1 \\
        \rowcolor{blue!15} \jepaname & ViT-B/16 & 400 & \cellcolor{red!20}75.7 & 57.1 & 64.8 & 94.6 & 78.8 & 90.4 & 82.9 & 88.3 & 79.1 & 75.7 \\
        \rowcolor{blue!15} \jepaname & ViT-L/14 & 400 & \cellcolor{red!20}\textbf{76.5} & 56.6 & 66.2 & 94.1 & 71.9 & 90.2 & 83.3 & 89.2 & 78.5 & 77.0 \\
        \hline
        \multicolumn{3}{l}{\gray{w/o heuristics - LP 100 epochs}}  & \cellcolor{white} & & & & & & & &\\
        LeJEPA$^*$ & ViT-L/14 & 100 & \textbf{78.3} & 57.0 & 57.3 & \textbf{96.5} & 83.7 & 91.2 & 82.1 & \textbf{89.7} & 79.5 & 75.6\\
        \rowcolor{blue!15} \jepaname & ViT-L/14 & 100 & \cellcolor{red!20}76.3 & \textbf{57.8} & \textbf{66.8} & 95.9 & \textbf{84.2} & \textbf{92.3} & \textbf{83.9} & 88.7 & \textbf{80.7} & 75.6 \\
    \hline
    \end{tabular}
\end{table*}

%% file: tables/lp-ood-test.tex
\setlength{\tabcolsep}{2.9mm}
\begin{table*}
    \centering
    \caption{Linear probe performance on the OOD downstream datasets. Similar to the observation in Table~\ref{tab:in-domain}, \jepaname has a better capability on handling OOD tasks/data. With a larger training set, \emph{i.e.}, ImageNet22K, \jepaname can achieve a comparable accuracy to DINOv2 by using \textbf{0.1x} of the training data. The \textbf{best} and \underline{second best} accuracy values are highlighted. Retina. and OrganA. stand for RetinaMNIST and OrganAMNIST.}
    \label{tab:lp-ood}
    \begin{tabular}{lccccccc|c}
    \hline
        Methods & Backbone & DTD & Galaxy10 & AID & ChestXRay & Retina. & OrganA. & Avg.\\ 
         \hline
        \gray{w/ heuristics} & & & & & & &\\
        MoCoV3 & ViT-B/16 & 73.72 & 73.06 & 90.20 & 23.89 & \underline{64.50} & 91.37 & 69.46\\
        DINO & ViT-B/16 & 74.26 & 72.77 & \underline{91.52} & 24.63 & 62.50 & 91.70 & 69.56\\
        data2vec & ViT-L/16 & 69.68 & 65.73 & 85.98 & 22.45 & 63.75 & 89.04 & 66.10\\
        iBOT & ViT-B/16 & 74.10 & 71.65 & \textbf{91.73} & \underline{24.80} & 63.75 & 91.84 & 69.64\\
        iBOT & ViT-L/16 & 75.27 & 72.66 & 91.24 & 23.94 & 63.00 & 90.82 & 69.49\\
        I-JEPA & ViT-H/14 & 69.89 & 71.31 & 88.68 & 23.46 & \textbf{65.75} & 92.19 & 68.55\\
        \gray{w/o heuristics} & & & & & & &\\
        MAE & ViT-L/16 & 72.77 & 71.98 & 86.42 & 22.87 & 63.00 & 90.06 & 67.85\\
        \rowcolor{blue!15}\jepaname & ViT-B/16 & \underline{75.69} & \underline{74.01} & 90.91 & \textbf{24.88} & 62.25 & \textbf{93.40} & \underline{70.19}\\
        \rowcolor{blue!15}\jepaname & ViT-L/14 & \textbf{76.54} & \textbf{76.32} & 90.12 & 23.64 & 64.25 & \underline{92.93} & \textbf{70.63}\\ \hline
        \gray{large scale datasets} & & & & & & &\\
        DINOv2-LVD142M & ViT-L/14 & \textbf{82.23} & 76.72 & \textbf{94.27} & 23.83 & \textbf{69.50} & 91.01 & 72.93 \\
        \rowcolor{blue!15}\jepaname-Inet22K & ViT-L/14 & 80.74 & \textbf{79.82} & 92.81 & \textbf{24.46} & 66.50 &  \textbf{93.33} & \textbf{72.94} \\
    \hline
\end{tabular}
\end{table*}

%% file: tables/ft-test.tex
\setlength{\tabcolsep}{0.7mm}
\begin{table}
    \centering
    \caption{Evaluating transfer learning capability. We fine-tune the pretrained \jepaname on five datasets and report the top-1 accuracy. The result indicates that \jepaname has a better transfer learning capability than DINO. The backbone is ViT-B/16, the accuracy on Galaxy10 is reproduced, the others values of supervised learning (Sup.)~\cite{sup} and DINO~\cite{dino} are from the orignal paper.}
    \label{tab:ft}
    \begin{tabular}{lccccc}
    \hline
         & CIFAR10 & CIFAR100 & Flowers & Inet1K & Galaxy10\\
    \hline
        Sup. & 99.0 & 89.5 & 98.5 & 81.5 & - \\
        DINO & 99.1 & 91.7 & 98.8 & 82.8 & 86.6 \\
        \rowcolor{blue!15}\jepaname  & \textbf{99.2} & \textbf{91.8} & \textbf{99.0} & \textbf{83.0} & \textbf{87.0}\\
    \hline
    \end{tabular}
\end{table}

%% file: tables/seg.tex
\setlength{\tabcolsep}{1.3mm}
\begin{table}[t]
    \centering
    \caption{Evaluation on dense instance prediction. \jepaname can produce a good result but the performance gap to the best, \emph{e.g.,} MoCoV3, is not negligible. The backbone is ViT-B/16, the metric is mIoU, and the values are reproduced.}
    \label{tab:seg}
    \begin{tabular}{c|cccc >{\columncolor{blue!15}}c}
    \hline
        Methods & MoCoV3 & DINO & data2vec & MAE & \jepaname \\
        ADE20K & \textbf{31.69} & 29.40 & 21.99 & 23.60 & 30.16\\
    \hline
    \end{tabular}
    \vspace{-2mm}
\end{table}

%% file: tables/generation.tex
\setlength{\tabcolsep}{1.2mm}
\begin{table}
    \centering
    \caption{Image generation results. Following iREPA~\cite{irepa}, we train SiT-B/2 for 100K steps with the guidance of DINO and \jepaname. The evaluation follows the standard 50K generation w/o CFG~\cite{adm}. \jepaname achieves better results across all metrics.}
    \label{tab:gen}
    \begin{tabular}{lccccc}
    \hline
        Methods & backbone & IS$\uparrow$ & gFID$\downarrow$ & Precision$\uparrow$ & Recall$\uparrow$\\
        DINO & ViT-B/16 & 33.47 & 41.15 & 50.51 & 60.70 \\
        \rowcolor{blue!15}\jepaname & ViT-B/16 & \textbf{33.48} & \textbf{40.36} & \textbf{51.38} & \textbf{61.26} \\
    \hline
    \end{tabular}
    \vspace{-5mm}
\end{table}

%% file: sec/conclusion.tex
\section{Conclusion}
This paper proposes \jepaname, a self-supervised learning method that does not rely on heuristics for training stability. We present its effectiveness on model scaling, training stability, and training efficiency. In addition, we show that \jepaname and its alike method is more robust to low-quality datasets than DINO, which is helpful in real-world applications. Last, we conduct extensive experiments to evaluate its performance in important aspects of a foundation model training method. It is intriguing that \jepaname has a stronger performance on OOD data and transfer learning. With this potential, we hope this technical path can enhance the usefulness of the foundation models.

%% file: sec/supp.tex
\section{Implementation details}

\subsection{Training details}
\sethlcolor{red!15}
\textbf{\hl{Pretraining on ImageNet-1K.}} We pretrain two model variants on ImageNet-1K~\cite{imagenet} (1.28M training images):
\textbf{VISReg-B} (ViT-B/16, 86M parameters) and \textbf{VISReg-L} (ViT-L/14, 304M parameters).
Both models are trained from scratch using the VISReg regularization objective and timm for backbones.

We adopt DINO-style multi-crop augmentation~\cite{dino}: each image produces
$N_g{=}4$ global crops ($224{\times}224$, scale $[0.3, 1.0]$) and $N_l{=}6$ local crops
($96{\times}96$ for ViT-B, $98{\times}98$ for ViT-L, scale $[0.05, 0.3]$),
yielding 10 views per image.
Augmentations include random horizontal flip, color jitter ($p{=}0.8$),
random grayscale ($p{=}0.2$), Gaussian blur ($p{=}0.5$), and random solarize ($p{=}0.2$).

We use AdamW with weight decay $5{\times}10^{-2}$ and bfloat16 mixed precision.
The learning rate follows a linear warmup over 5 epochs
then cosine annealing to $\mathrm{lr}_{\max}/1000$.
Projections are produced by a 3-layer MLP ($2048 \to 2048 \to d_p$)
with batch normalization and GELU activations,
applied to the concatenated CLS tokens from the last two backbone layers.

\textbf{VISReg-B} uses learning rate $9{\times}10^{-4}$, $\lambda{=}0.9$,
projection dimension $d_p{=}256$, $K{=}2048$ random projections for VISReg,
and per-GPU batch size 16 (effective batch size 512 across 32 GPUs).
\textbf{VISReg-L} uses learning rate $8{\times}10^{-4}$, $\lambda{=}0.7$,
$d_p{=}384$, $K{=}4096$ random projections,
and per-GPU batch size 16 (effective batch size 512 across 32 GPUs).
Both models are trained for 400 epochs on 32 NVIDIA H100 80GB GPUs (4 nodes $\times$ 8 GPUs)
using HuggingFace Accelerate for distributed training,
requiring approximately 1{,}120 and 2{,}060 GPU-hours for ViT-B and ViT-L, respectively.

\sethlcolor{blue!15}
\textbf{\hl{Pretraining on ImageNet-22K.}}
We additionally pretrain VISReg on ImageNet-22K~\cite{imagenet} (14.2M images) with ViT-L/14
for 100 epochs on 16 NVIDIA H100 80GB GPUs (4 nodes $\times$ 4 GPUs).
The multi-crop strategy uses $N_g{=}2$ global crops and $N_l{=}8$ local crops ($98{\times}98$),
still yielding 10 views per image.
We use per-GPU batch size 64 (effective batch size 1{,}024),
learning rate $8{\times}10^{-4}$, $\lambda{=}0.8$,
$d_p{=}384$, and $K{=}4096$ random projections.
All other settings (optimizer, scheduler, projector architecture)
follow the ImageNet-1K configuration.
Training requires approximately 2{,}720 GPU-hours.

\subsection{Testing details}
\sethlcolor{green!15}
\textbf{\hl{Downstream classification.}}
We evaluate pretrained representations on 8 classification benchmarks following the
DINOv2 linear evaluation protocol~\cite{dinov2}:
DTD~\cite{dtd}, FGVC-Aircraft~\cite{aircrafts},
Stanford Cars~\cite{cars}, CIFAR-10, CIFAR-100~\cite{cifar},
Oxford Flowers-102~\cite{flowers}, Food-101~\cite{food},
and Oxford-IIIT Pets~\cite{pets}.
We additionally evaluate on 6 out-of-distribution benchmarks:
DTD, Galaxy10~\cite{galaxy10}, AID~\cite{aid},
NIH ChestX-ray~\cite{chestxray}, RetinaMNIST, and OrganAMNIST~\cite{organamnist}. The details of each dataset can be found under \ref{supp/dataset}.

For all classification tasks, we freeze the pretrained encoder and extract features by
concatenating the CLS tokens from the last 4 transformer layers,
yielding a feature vector of dimension $4 \times d_{\mathrm{embed}}$
(3{,}072 for ViT-B, 4{,}096 for ViT-L).
A linear classifier with SyncBatchNorm is trained on top of these frozen features
using SGD with momentum 0.9, no weight decay, and cosine annealing
for 10 epochs with batch size 32.
We perform a grid search over 13 base learning rates
scaled by the linear scaling rule (effective batch size / 256),
and report the best test accuracy.
All images are resized to $224{\times}224$ with standard ImageNet normalization.

\sethlcolor{cyan!15}
\textbf{\hl{ImageNet-1K linear probe.}}
For ImageNet-1K linear evaluation, we follow the same frozen-feature protocol
but with a dedicated multi-head implementation for efficiency.
A multi-head linear classifier with shared SyncBatchNorm
trains 10 independent heads in parallel (one per learning rate),
for 100 epochs using SGD with momentum 0.9, no weight decay,
and per-step cosine annealing.
Training uses bfloat16 mixed precision on 8 GPUs with per-GPU batch size 32
(effective batch size 256).
Standard ImageNet evaluation preprocessing is applied:
random resized crop to $224{\times}224$ for training,
resize to 256 then center crop to 224 for validation.
The best accuracy across all heads on the 50K validation set is reported.

\sethlcolor{yellow!15}
\textbf{\hl{Semantic segmentation.}}
We evaluate on ADE20K~\cite{ade20k} (150 classes) using a linear segmentation probe.
A single $1{\times}1$ convolution with SyncBatchNorm is trained on frozen patch features
from the last transformer layer.
Training uses AdamW with learning rate $2{\times}10^{-3}$,
polynomial LR decay (power 0.9), batch size 16, and image size $518{\times}518$
(for patch-14 models) or $512{\times}512$ (for patch-16 models)
for 40 epochs.
We report mean intersection-over-union (mIoU) on the validation set.

\subsection{Datasets}
\label{supp/dataset}

We evaluate our pre-trained models on a diverse set of 15 datasets spanning multiple domains, tasks, and difficulty levels. The following sections describe each dataset in detail.

\sethlcolor{red!15}
\textbf{\hl{ImageNet-1k}}~\cite{imagenet} is a large-scale object recognition dataset containing 1,281,167 training images and 50,000 validation images across 1,000 classes, representing diverse natural objects, animals, and scenes from the natural world.

\sethlcolor{blue!15}
\textbf{\hl{CIFAR-10}}~\cite{cifar} is a 10-class object classification dataset with 50,000 training and 10,000 test images at $32\times32$ resolution, covering categories like airplanes, cars, birds, cats, and other common objects. 

\sethlcolor{cyan!15}
\textbf{\hl{CIFAR-100}}~\cite{cifar} is a fine-grained object classification dataset with 50,000 training and 10,000 test images at $32\times32$ resolution, containing 100 classes organized into 20 supercategories including various vehicles, animals, and household items.

\sethlcolor{green!15}
\textbf{\hl{Stanford Cars}}~\cite{cars} is a fine-grained vehicle classification dataset with 8,144 training and 8,041 test images covering 196 car models from 98 manufacturers, spanning decades of automotive design from 1950 to 2012. 

\sethlcolor{magenta!15}
\textbf{\hl{Galaxy10}}~\cite{galaxy10} is an astronomical image classification dataset with 17,736 images classifying galaxies into 10 morphological categories (disturbed, merging, spiral, elliptical, etc.) from the DECaLS survey. 

\sethlcolor{orange!15}
\textbf{\hl{Food-101}}~\cite{food} is a food classification dataset with 75,750 training and 25,250 validation images covering 101 food categories including dishes like pizza, sushi, hamburger, and various international cuisines. 

\sethlcolor{purple!15}
\textbf{\hl{Oxford-IIIT Pets}}~\cite{pets} is a pet breed classification dataset with 3,680 training and 3,669 test images covering 37 cat and dog breeds, requiring fine-grained distinction between similar-looking breeds. 

\sethlcolor{teal!15}
\textbf{\hl{NIH Chest X-ray}}~\cite{chestxray} is a multi-label medical image classification dataset comprising 112,120 total images, where each chest radiograph may contain multiple pathology labels from 14 disease categories including pneumonia, cardiomegaly, and pleural effusion. 

\sethlcolor{lime}
\textbf{\hl{RetinaMNIST}}~\cite{retinamnist} is a medical image classification dataset with 1,080 training, 120 validation, and 400 test retinal fundus images classifying diabetic retinopathy into 5 severity grades. 

\sethlcolor{olive!15}
\textbf{\hl{OrganAMNIST}}~\cite{organamnist} is a medical image classification dataset with 34,581 training, 6,491 validation, and 17,778 test CT axial slices classifying 11 body organ types including liver, kidney, spleen, and heart. 

\sethlcolor{pink}
\textbf{\hl{Oxford Flowers 102}}~\cite{flowers} is a fine-grained plant classification dataset with 1,020 training, 1,020 validation, and 6,149 test images covering 102 flower species with 40-258 images per class. 

\sethlcolor{brown!15}
\textbf{\hl{Describable Textures (DTD)}}~\cite{dtd} is a texture classification dataset with 1,880 training, 1,880 validation, and 1,880 test images across 47 texture categories (e.g., braided, dotted, fibrous) following a 10-fold cross-validation protocol.

\sethlcolor{lightgray}
\textbf{\hl{FGVC-Aircraft}}~\cite{aircrafts} is a fine-grained aircraft classification dataset with 6,667 trainval and 3,333 test images covering 100 aircraft variants from the FGVC-Aircraft 2013b benchmark. 

\sethlcolor{violet!15}
\textbf{\hl{AID}}~\cite{aid} is a remote sensing scene classification dataset with 10,000 images across 30 aerial scene categories including airports, beaches, forests, and urban areas, using a 10\%/90\% train/test split for SSL evaluation. 

\sethlcolor{yellow}
\textbf{\hl{ADE20K}}~\cite{ade20k} is a semantic segmentation dataset with 20,210 training and 2,000 validation images, containing pixel-level annotations for 150 semantic classes including objects, parts, and materials across indoor and outdoor scenes. 

\section{Additional ablations}
\label{ablation:visreg}
Our additional ablations focus on VISReg design, including the necessity of scale, shape, center loss, the necessity of applying gradient detachment between scale and shape loss, and the effect of the loss weight on each component. 
We include a long-tailed dataset (ImageNet-LT), a low-rank dataset (Galaxy10), and a normal dataset (Imagenette) to cover a wider range of application scenarios.
All experiments use a ViT-S/8 backbone at $128{\times}128$ resolution with 4 augmented views,
learning rate $10^{-3}$, per-GPU batch size 32 across 8 GPUs (effective batch 256),
$\lambda=0.6$, projection dimension 256, and $K{=}4096$ random projections.
ImageNet-LT and Galaxy10 train for 400 epochs; Imagenette for 800.

\subsection{Effect of decoupled components in training.}
First, we knocked out each contribution of scale, shape, and center in the training to understand the effectiveness of each part. Since the result on ImageNette has shown a clear pattern, ImageNet-LT and Galaxy are not included. Table~\ref{tab:visreg-knockout} shows that both scale and shape loss significantly impact the learning process: 1) Without scale loss, there is a 71.02\% decrease in accuracy; Without shape loss, there is a 58.4\% decrease in accuracy. As for the center loss, the accuracy difference is 0.41\% in the final accuracy, but importantly, it increases the convergence speed. Hence, all three components are necessary.
\input{tables/visreg-comp-ablation}

Second, we check the usefulness of applying detachment between scale loss and shape loss. The general observation from Table~\ref{tab:visreg-detach} is that, despite the minor improvement, detachment helps the model achieve a higher performance across all three datasets. Therefore, we choose to use it across all the experiments.

\subsection{Effect of decoupled components in different training set scenarios.}
We ablate the ratio between the three DSSO loss components, \emph{i.e.}, scale, shape, and center, while
keeping the total weight constant ($\lambda_\text{scale} + \lambda_\text{shape} + \lambda_\text{center} = 3$)
so that $\lambda$ alone controls the overall regularization magnitude.

The shape component is
the most impactful of the three DSSO objectives.
On ImageNet-LT and Galaxy10, shifting weight toward shape monotonically improves accuracy,
with shape 4:1 outperforming the equal baseline by +3.2\% and +1.3\%, respectively.
Conversely, emphasizing scale or center consistently degrades performance,
with scale 4:1 producing the largest drops (-4.6\% on ImageNet-LT, -2.7\% on Galaxy10).
This suggests that a higher shape regularization helps the learning on low-quality datasets. However, the result on Imagenette shows that the default setting is the best choice. Other imbalanced ratio across three factors largely reduces the learning effectiveness.
\input{tables/shape-scale-center-abation}

\section{Visualizations}
\subsection{VISReg loss indicates the performance}
Strong correlation between loss and online probe accuracy is a important advantage of the theorem proposed by LeJEPA~\cite{lejepa}. We calculate the Pearson correlation between loss and online probe accuracy of the ViT-L training on ImageNet1K, as shown in Figure~\ref{fig:loss-acc-corr}. The pronounced -0.996 correlation show the loss curve can be used to reflect the learning curve of the model.

\input{figures/loss-acc-corr}

\subsection{Further comparison with DINOv1 on image and video.}
\paragraph{PCA Feature Visualization.}
To qualitatively compare the learned representations, we visualize patch-level features from different ViT encoders using PCA coloring.
For each input image, we extract the spatial patch token features from the last layer of the encoder, yielding a feature map of shape $H_p \times W_p \times C$, where $H_p$ and $W_p$ are the patch grid dimensions and $C$ is the embedding dimension.
We flatten this to an $N \times C$ matrix ($N = H_p \times W_p$) and apply PCA to reduce it to three components, which are then interpreted as RGB channels.
Each component is independently normalized to $[0,1]$ via min--max scaling, and the resulting $H_p \times W_p \times 3$ map is bilinearly upsampled to the original image resolution for display.
Since PCA components are determined only up to permutation and sign, direct comparison between models requires alignment.
We compute the $3 \times 3$ Pearson correlation matrix between the PCA components of a reference model (\emph{i.e.}, DINO ViT-B/16~\cite{dino}) and the target model (\emph{i.e.}, VISReg ViT-B/16), then solve the optimal assignment using the Hungarian algorithm on $-\lvert\mathrm{corr}\rvert$.
The matched target components are reordered accordingly and flipped in sign where the correlation is negative, ensuring consistent color semantics across models.
For video visualizations, PCA is fit jointly on the patch features from all frames to ensure temporal consistency of the color mapping. Both Figure~\ref{fig:image-vis} and Figure~\ref{fig:video-vis} indicates that VISReg helps model learn more granular details than DINO.
\input{figures/pca-visual-video}
\input{figures/PCA-visual-image}

%% file: tables/visreg-comp-ablation.tex
\begin{table}[t]
    \centering
    \begin{minipage}[t]{0.49\textwidth}
        \centering
        \caption{VISReg component ablation on ImageNette. Scale loss and shape loss are necessary for convergence. Center loss is helpful for faster and better learning.}
        \label{tab:visreg-knockout}
        \smallskip
        \begin{tabular}{lccc c}
        \toprule
        Config & Scale & Shape & Center & Best Accuracy \\
        \midrule
        No scale  & \xmark & \cmark & \cmark & 20.80 \\
        No shape  & \cmark & \xmark & \cmark & 33.42 \\
        No center & \cmark & \cmark & \xmark & 91.41 \\
        VISReg    & \cmark & \cmark & \cmark & 91.82 \\
        \bottomrule
        \end{tabular}
    \end{minipage}%
    \hfill
    \begin{minipage}[t]{0.49\textwidth}
        \centering
        \caption{Necessity of detach ablation. Fully decoupling scale and shape loss helps the learning in all three tasks.}
        \label{tab:visreg-detach}
        \begin{tabular}[t]{c|ccc}
        \hline
        ViT-S/8 & Imagentte & ImageNet-LT & Galaxy10 \\ \hline
        w/o detach & 90.62 & 31.59 & 78.91 \\
        w/ detach & \textbf{90.89} & \textbf{31.94} & \textbf{79.44} \\
        \hline
        \end{tabular}
    \end{minipage}
\end{table}

%% file: tables/shape-scale-center-abation.tex
\setlength{\tabcolsep}{2.5mm}
\begin{table}[t]
\centering
\caption{VISReg weight ratio ablation. A higher regularization on shape is helpful for low-quality datasets but not for high-quality datasets. \textbf{Bold} marks the best result per dataset. }
\label{tab:dsso-ratio-ablation}
\smallskip
\begin{tabular}{l|ccc|ccc}
\hline
Config & $\lambda_{scale}$ & $\lambda_{shape}$ & $\lambda_{center}$ & ImageNet-LT & Galaxy10 & Imagenette \\
\hline
Baseline & 1.0  & 1.0  & 1.0  & 31.94 & 79.44 & \textbf{91.82} \\
Scale 2:1        & 1.5  & 0.75 & 0.75 & 30.16 & 78.22 & 91.50 \\
Shape 2:1        & 0.75 & 1.5  & 0.75 & 34.25 & 80.03 & 91.31 \\
Center 2:1       & 0.75 & 0.75 & 1.5  & 31.60 & 78.91 & 90.94 \\
Scale 4:1        & 2.0  & 0.5  & 0.5  & 27.35 & 76.71 & 90.77 \\
Shape 4:1        & 0.5  & 2.0  & 0.5  & \textbf{35.14} & \textbf{80.76} & 89.87 \\
Center 4:1       & 0.5  & 0.5  & 2.0  & 29.48 & 74.95 & 90.36 \\
\hline
\end{tabular}
\end{table}

%% file: figures/loss-acc-corr.tex
\begin{figure}[t]
    \centering
    \includegraphics[width=0.6\linewidth]{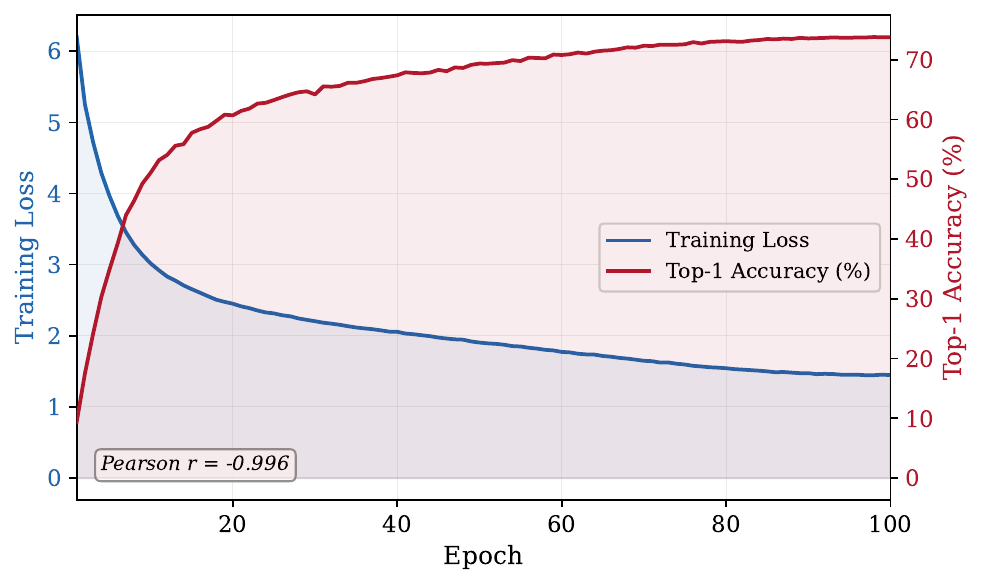}
    \caption{Pearson correlation between loss curve and online accuracy curve. The data is from the ViT-L/14 training on ImageNet1K for 100 epochs. The -0.996 correlation strongly suggests that loss curve can reflect the learning curve of the model.}
    \label{fig:loss-acc-corr}
\end{figure}

%% file: figures/pca-visual-video.tex
\begin{figure}[t]
    \centering
    \includegraphics[width=\linewidth]{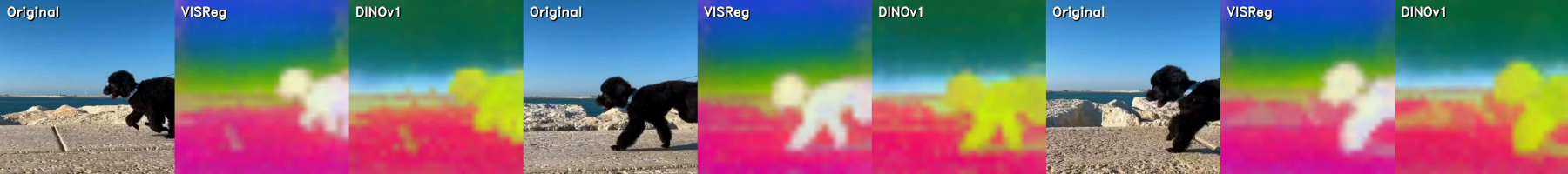}
    \caption{PCA visualization of three video frames. VISReg can learn better concepts and details than DINOv1.}
    \label{fig:video-vis}
\end{figure}

%% file: figures/PCA-visual-image.tex
\begin{figure}[h]
    \centering
    \includegraphics[width=\linewidth]{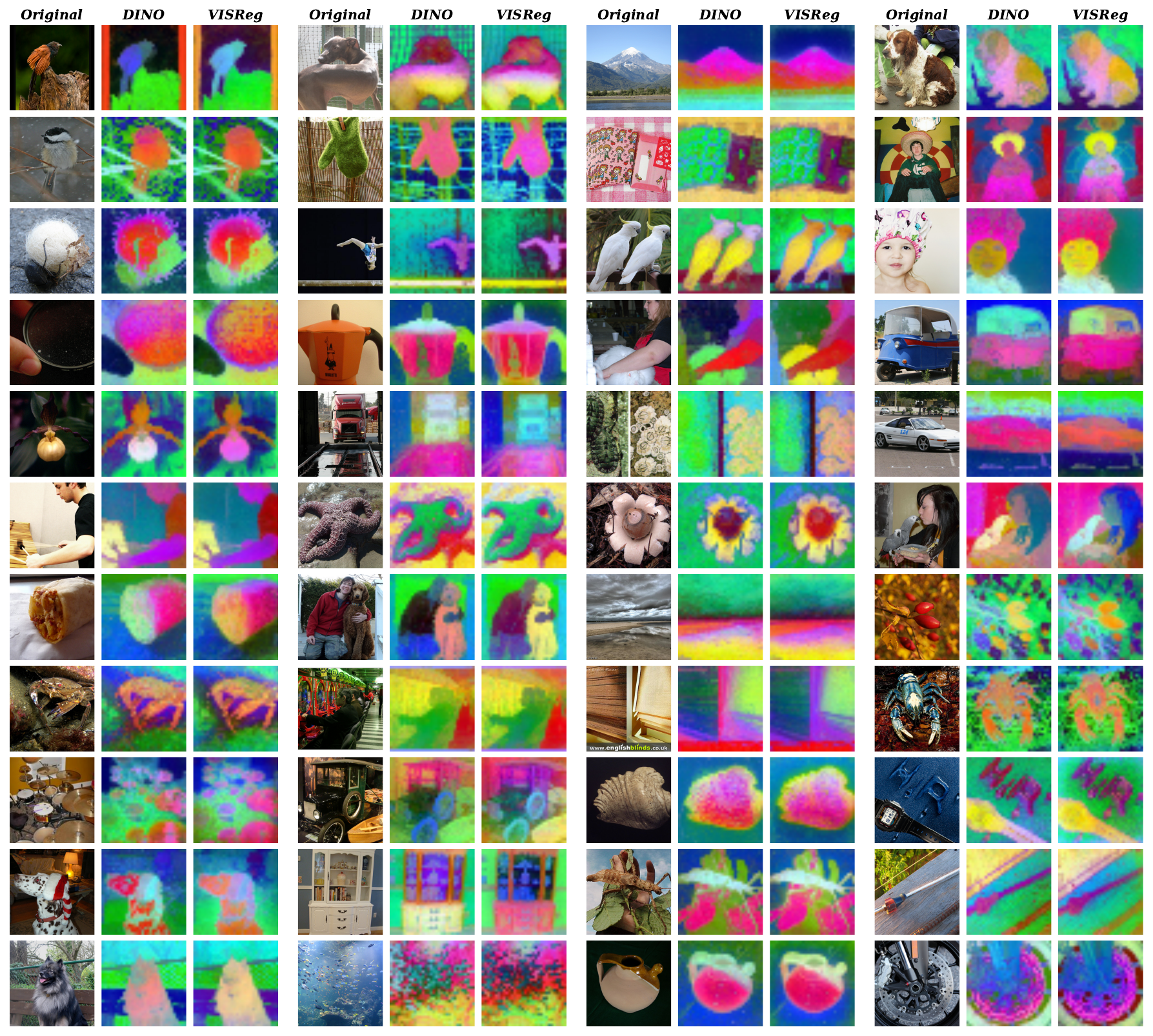}
    \caption{PCA visualization of the ImageNet1K images. Similarly to Figure~\ref{fig:video-vis}, VISReg can learn better concepts and details than DINOv1.}
    \label{fig:image-vis}
\end{figure}